# Post-Processing of Discovered Association Rules Using Ontologies

Claudia Marinica, Fabrice Guillet and Henri Briand
*LINA – Ecole polytechnique de l'Université de Nantes, France*
*{claudia.marinica, fabrice.guillet, henri.briand}@univ-nantes.fr*

**Abstract**

*In Data Mining, the usefulness of association rules is strongly limited by the huge amount of delivered rules. In this paper we propose a new approach to prune and filter discovered rules. Using Domain Ontologies, we strengthen the integration of user knowledge in the post-processing task. Furthermore, an interactive and iterative framework is designed to assist the user along the analyzing task. On the one hand, we represent user domain knowledge using a Domain Ontology over database. On the other hand, a novel technique is suggested to prune and to filter discovered rules. The proposed framework was applied successfully over the client database provided by Nantes Habitat[1].*

## 1. Introduction

The technique of mining association rules, introduced in [2], is considered as one of the most relevant tasks in Knowledge Discovery in Databases [10]. It aims to discover, among sets of items in transaction databases, implicative tendencies that can be revealed as being valuable information.

An association rule is described as the implication $X \rightarrow Y$ where $X$ and $Y$ are sets of items and $X \cap Y = \phi$. The strength of association rule mining rests in its ability to deliver interesting discovered knowledge that exists in data. Unfortunately, due to high dimensionality of massive data, this strength becomes its main weakness when analyzing the mining result. The huge number of discovered rules makes very difficult for a decision maker to manually outline the interesting rules. Thus, it is crucial to help the decision maker with an efficient reduction of the number of rules.

To overcome this drawback, the post-processing task was proposed to improve the selection of discovered rules. Different complementary post-processing methods may be used, like pruning, summarizing, grouping or visualization [4]. The pruning phase consists of removing uninteresting or redundant rules. In the summarizing phase summaries of rules are generated. Groups of rules are produces in the grouping phase; meanwhile the visualization phase is useful to have a better presentation.

However, most of existing post-processing methods are generally based on statistical information on database. Since rule interestingness strongly depends on user knowledge and goals these methods are not efficient enough. For instance, if the user looks for unexpected rules, all the already known rules should be pruned. Or, if the user wants to focus on specific schemas of rules, only this subset of rules should be selected.

This paper proposes a new approach to prune and filter discovered rules. Using Domain Ontologies, we strengthen the integration of user knowledge in the post-processing task. Furthermore, an interactive and iterative framework is designed to assist the user along the analyzing task. On the one hand, we represent user domain knowledge using a Domain Ontology over database. On the other hand, a novel technique is suggested to prune and to filter discovered rules. User expectations are described by the notion of Rule Schema and rule operators are proposed to guide user actions. Ontologies will offer a powerful representation of user knowledge, and rule schemas and rule operators a more expressive representation of user expectations in terms of rules.

The paper is structured as follows. Section 2 describes the research domain and reviews related works. Section 3 explains the proposed framework and its elements. Section 4 is devoted to the elements of the framework: the mining process, the user knowledge and the post-processing step. Finally, section 5 presents conclusions and shows directions for future research.

## 2. Related Works

### 2.1. Post-Processing Techniques

Several approaches, integrating user knowledge, to solve the problem of huge number of discovered rules have been proposed. As early as 1994, in the KEFIR system [16], the key finding and deviation notions were suggested. Grouped in findings, deviations represent the difference between the actual value and the expected value.

---

[1] We would like to thank Nantes Habitat, the Public Housing Unit in Nantes, France, and more specially Ms. Christelle Le Bouter for supporting this work.

Later, Klemettinen et al. proposed templates [12] to describe the form of interesting rules (inclusive templates), and those of not interesting rules (restrictive templates). Other approaches proposed to use a rule-like formalism to express user expectations, and the discovered rules are pruned/summarized comparing them to user expectations ([17], [15], [13]).

Toivonen et al. proposed in [19] a novel technique for rule pruning and grouping based on *rule covers*. The notion of *rule cover* defines the subset of a rule set describing the same transaction row. Thus, the authors define the pruning action as the reduction of a rule set to its rule cover.

The notion of subsumed rules, discussed in [6], describes a set of rules having the same consequent and several additional conditions in the antecedent with respect to another rule. Bayardo Jr. et al. proposed a new pruning measure described as the difference between the confidences of the two rules, called Minimum Improvement. A rule is pruned if this measure is less than a pre-specified threshold, so the subsumed rule does not bring a lot of information comparing to the other rule.

In the Web domain, the paper [1] presents a framework for building behavioral profiles of individual users. Considering a set of discovered rules for each client, the authors propose an iterative rule validation process based on several operators, including rule grouping, filtering, browsing, and redundant rule elimination.

Another related approach is proposed by An et al. in [3] where the authors introduce domain knowledge in order to prune and summarize discovered rules. The first algorithm proposed use a data taxonomy, proposed by user, in order to describe the semantic distance between rules and to group rules. The second algorithm allows to group discovered rules sharing at least an item in the antecedent and in the consequent.

An original proposition was made in [8] with the exploitation of the directed hypergraphs in order to prune singleton consequent rules. Thus, the discovered rules are represented in a directed hypergraph called, after being pruned of cycles, Association Rules Network (ARN).

In 2007, a new methodology was proposed in [5] to prune and organize rules with the same consequent. The authors suggest transforming the database in an association-rulebase in order to extract second level association rules. Called *metarules*, the extracted rules $r1 \rightarrow r2$ express relations between the two association rules and help on pruning/grouping discovered rules.

## 2.2. Ontologies and Data Mining

Ontologies, introduced in data mining for the first time in early 2000, can be used in several ways [14]: Domain and Background Knowledge Ontologies, Ontologies for Data Mining Process, or Metadata Ontologies. Background Knowledge Ontologies organize domain knowledge and play important roles at several levels of knowledge discovery process. Ontologies for Data Mining Process codify mining process description and choose the most appropriate task according to the given problem; meanwhile, Metadata Ontologies describe the construction process of items.

In this study, we are interested in Domain and Background Knowledge Ontologies and we will present past studies related to them. The first idea of using Domain Ontologies was introduced by Srikant and Agrawal with the concept of Generalized Association Rules [18]. The authors proposed taxonomies of mined data (an is-a hierarchy) in order to generalize/specify rules.

In [7], and developed in [9], it is suggested that an ontology of background knowledge can benefit all the phases of a KDD cycle described in CRISP-DM. The role of ontologies is based on the given mining task and method, and on data characteristics. From business understanding to deployment, the authors delivered a complete example of using ontologies in a cardiovascular risk domain.

Related to Generalized Association Rules, the notion of *raising* was exposed in [20]. *Raising* is the operation of generalizing rules (making rules more abstract) in order to increase support in keeping confidence high enough. This allows for strong rules to be discovered and also to obtain sufficient support for rules that, before *raising*, would not have minimum support due to the particular items they referred to. The difference with Generalized Association Rules is that this solution proposes to use a specific level for *raising* and mining.

## 3. The Framework

The new approach defines a new formal environment to prune and group discovered associations integrating knowledge into specific mining process of association rules. It is composed of three main parts (as shown in Figure 1).

Firstly, a basic mining process is applied over data extracting a set of association rules. Secondly, the knowledge base allows formalizing user knowledge and goals. Domain knowledge allows a general view over user knowledge in database domain, and user expectations express user already knowledge over the discovered rules. Finally, the post-processing step consists in applying several operators (i.e. pruning) over user expectations in order to extract the interesting rules.

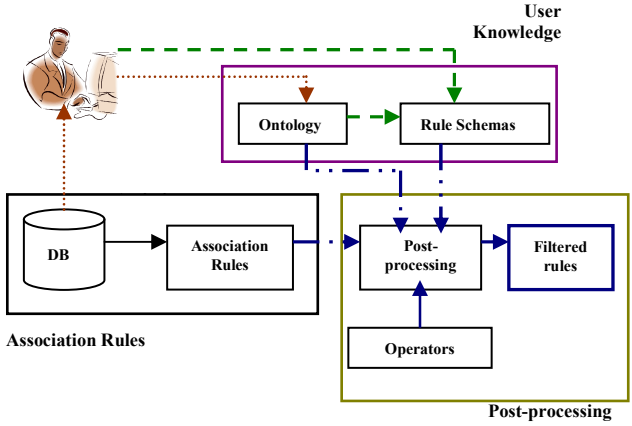

**Figure 1. Framework description**

The novelty of this approach resides in supervising the knowledge discovery process using different conceptual structures for user knowledge representation: one or several ontologies and several rule schemas.

### 3.1. Database and Association Rule Mining

The association rules mining techniques are applied over databases described as $D = \{I, T\}$. Let $I = \{I_1, I_2, \dots, I_p\}$ be the set of attributes (called items) and $T = \{t_1, t_2, \dots t_n\}$ be the transaction set. Each transaction $t_i = \{I_1, I_2, \dots, I_{mi}\}$ is a set of items, such as $t_i \subset I$ and each subset of items, $X$, is called itemset.

An association rule is an implication $X \rightarrow Y$, where $X$ and $Y$ are two itemsets and $X \cap Y = \phi$. This rule holds on $D$ with the confidence $c$ if $c\%$ of transactions in $T$ that contain $X$, also contain $Y$. The rule has support $s$ in transaction set $T$ if $s\%$ of transactions contain $X \cup Y$.

Since their early definition, association rules are mined using *Apriori* algorithm proposed for the first time in Agrawal et al., 1993.

### 3.2. User Knowledge

In association rule mining process, user knowledge can be divided into two main types: *domain knowledge*, mainly related to database items, and *user beliefs* expressing user expectations according to the discovered knowledge. In addition, we propose a third user-based element described by the actions that a user can realize among his/her different beliefs. Thus, the *operators* are introduced in order to guide the post-processing step. This element will be discussed in the following section.

An ontology is described as a formal explicit specification of a shared conceptualization for a domain of interest [11].

**Definition 1.** Formally, an ontology is a 3-tuple $O = \{C, R, H\}$. $C = \{C_1, C_2, \dots, C_o\}$ is a set of concepts and $R = \{R_1, R_2, \dots, R_r\}$ is a set of relations defined over concepts. $H$ is a directed acyclic graph (DAG) over concepts defined by the subsumption relation (*is-a* relation, $\leq$) between concepts. We say that $C_2$ *is-a* $C_1$, $C_2 \leq C_1$, if the concept $C_1$ subsumes the concept $C_2$.

In this approach, we propose a domain knowledge model based on ontologies connecting ontology concepts to a set of database items. Consequently, domain ontologies over database extend the notion of Generalized Association Rules based on taxonomies as a result of the generalization of the subsumption relation by the set $R$ of ontology relations. Besides, ontologies are used as filters over items, generating item families.

In this scenario, it is fundamental to connect the ontology to the database, each concept and each instance being instantiated in one/several items.

Considering that the set of concepts $C$ is defined as the union of three concepts subsets $C = C_0 \cup C_1 \cup C_2$:

- $C_0$ is defined as the set of leaf-concepts of the ontology connected in the easiest way to database.

$$C_0 = \{c_0 \in C \mid \nexists c' \in C, c' \leq c_0\}$$

In this manner, each concept from $C_0$ is associated to an item in the database.

$$f_0 : C_0 \rightarrow I$$
$$\forall c_0 \in C_0, i \in I, i = f_0(c_0)$$

- $C_1$ is described as the set of generalized concepts in the ontology. A generalized concept is connected to database through its subsumed concepts. That means that, recursively, only the leaf-concepts subsumed by a generalized concept contribute to its database connection.

$$f : C_1 \rightarrow 2^I$$
$$\forall c \in C_1, f(c) = \{i = f_0(c_0) \mid c_0 \in C_0, c_0 \leq c\}$$

- More generally, we propose the definition of ontology concepts by logical expressions defined over items, organized in the $C_2$ subset. In a first attempt, we base the description of the logical expression on the *OR* logical operator. Thus the defined concept associated could be connected to a disjunction of items.

$$f : C_2 \rightarrow 2^I, \forall c \in C_2$$
$$c \rightarrow E(c)$$
$$f(c) = \{f(c') \mid c' \in E(c)\}$$

To improve association rule selection, we propose a rule filtering model, called Rule Schemas. In other words, a rule schema describes, in a rule-like formalism, the user expectations in terms of interesting/obvious rules. As a result, Rule Schemas act as a rule grouping, defining rule families.

The base of Rule Schema formalism is the user representation model introduced by Liu et al. in [13] composed of: General Impressions, Reasonably Precise Concepts and Precise Knowledge. The proposed model is described using elements from an attribute taxonomy allowing an *is-a* organization of database attributes.

A Rule Schema is a semantic extension of the Liu model since it is described using concepts from the domain ontology. We propose to develop two of the three representations introduced in [13]: General Impressions and Reasonably Precise Concepts. Thus, rule schemas bring the complexity of ontologies in rule mining combining not only item constraints, but also ontology concept constraints.

**Definition 2.** A rule schema is defined as:

$$\langle X_1, X_2, \ldots, X_{s1} (\rightarrow) Y_1, Y_2, \ldots, Y_{s2} \rangle$$

where $X_i$ and $Y_j$ are ontology concepts and the implication "$\rightarrow$" is optional. In other words, we can note that the proposed formalism combines General Impressions and Reasonably Precise Concepts. Consequently, if we use the formalism as an implication, an *implicative rule schema* is defined extending the Reasonably Precise Concepts. Meanwhile, if we do not keep the implication, we define *non implicative rules schemas*, generalizing General Impressions.

For example, a rule schema $C_2, \overline{C_3} \rightarrow C_4$ corresponds to "all association rules whose condition verifies $C_2$ and doesn't verify the concept $C_3$, and whom conclusion verifies $C_4$".

### 3.3. Operations in Post-processing Step

The post-processing task that we design is based on operators applied over rule schemas allowing to user to perform several actions over the discovered rules. We propose two important operators: pruning and filtering association rules. The filtering operator is composed by three operators: conforming, unexpectedness and exception.

These four operators will be presented along this section. To this end, let us consider an *implicative rule schema* $RS_1 : \langle X \rightarrow Y \rangle$, a *non implicative rule schema* $RS_2 : \langle U, V \rangle$ and an association rule $AR_1 : A \rightarrow B$ where *X, Y, U, V* are ontology concepts, and *A, B* are itemsets.

The *pruning operator* allows to user to remove families of rules that he/she considers that are uninteresting. In a database, there exist, in most of cases, relations between items that we consider obvious or that we already know. Thus, it is not useful to find these relations among the discovered associations. The pruning operator applied over a rule schema, *P(RS)*, eliminates all association rules matching the rule schema. To extract all the rules matching a rule schema the conforming operator is used.

The *conforming operator* applied over a rule schema, *C(RS)*, proposes to confirm an implication or to find the implication between several concepts. As a result, rules matching all the elements of a non-implicative rule schema are filtered. For an implicative rule schema, the condition and the conclusion of the association rule should match those of the schema.

The rule $AR_1$ is selected by the operator $C(RS_1)$ if both the condition and the conclusion of the rule $AR_1$ respectively match the condition and the conclusion of $RS_1$. Translating this description into the ontological definition of concepts means that $AR_1$ is conforming to $RS_1$ if:

$$\exists i \in f(X), i \in A \text{ and } \exists i \in f(Y), i \in B$$

Similarly, rule $AR_1$ is filtered by $C(RS_2)$ if the condition and/or the conclusion of the rule $AR_1$ match the schema $RS_2$:

$$\forall i \in f(U), i \in A \cup B, \text{ and } \forall i \in f(V), i \in A \cup B,$$

The *unexpectedness operator, U(RS)*, with a higher interest for the user, proposes to filter a set of rules with a surprise effect for the user. This type of rules interests the user more than the conforming ones since a decision maker generally searches to discover new knowledge with regard to his/her prior knowledge.

Moreover, several types of unexpected rules can be filtered according to the rule schema: rules that do not confirm either or both the condition and the conclusion of a rule schema.

For instance, let us consider that the operator $U(RS_1)$ extracts the rule $AR_1$ which is *unexpected according to the condition* of the rule schema $RS_1$. This is possible if rule conclusion *B*, matches the schema conclusion *Y*, while the condition, *A*, is unexpected according to the schema condition *X*:

$$\forall i \in f(X), i \notin A \text{ and } \exists i \in f(Y), i \in B$$

In a similar way, the two other unexpectedness operator usability are defined.

Finally, the *exception operator* applied over $RS_1$, is defined only over implicative rule schemas and extracts conforming rules with respect to the following new implicative rule schema: $X \wedge Z \rightarrow \overline{Y}$, where Z is a set of items.

## 4. Case study

### 4.1. Database

In this study, we use a questionnaire database, about client satisfaction concerning accommodation, provided by Nantes Habitat. The database consists in an annual study (since 2003) performed by Nantes Habitat on 1500 out of a total of 50000 clients.

**Table 1. Database extract**

| q1 | q2 | Q3 | q4 | q5 | q6 | q7 | Q8 | q9 | q10 |
|----|----|----|----|----|----|----|----|----|-----|
| 4  | 99 | 2  | 1  | 3  | 1  | 2  | 2  | 1  | 1   |
| 2  | 1  | 1  | 1  | 3  | 4  | 2  | 1  | 1  | 1   |
| 1  | 1  | 1  | 1  | 1  | 1  | 3  | 4  | 2  | 2   |

The questionnaire consists of 67 questions; each question may be answered with one of 6 possible answers. The first four answers express the degree of satisfaction: *"quite"=1, "rather"=2, "rather not"=3, "not at all"=4*. One answer is for the non applicable cases - *95/96* (i.e. questions concerning elevators – not all buildings are equipped with elevators) and the last one is for the cases when the client doesn't know the answer - *99*.

For instance, the item $q1=1$ is described by the question $q1=$*"Is your district transport practical?"* with the answer *1="quite"*. A database extract of the first 10 questions with the first 3 recordings is presented in Table 1.

To extract the association rules we resort to *Weka*[2] software, and we fix a minimum support of 2% and a maximum support of 30% to target the most interesting rules, and a minimum confidence of 80%. The *Apriori* algorithm extracts *82,159* association rules.

For example, the following association rule describes the relationship between questions q2, q3, q47 and question q70. Thus, if the clients are satisfied by the *access to the city center* ($q2$), *the shopping facilities* ($q3$) and *the apartment ventilation* ($q47$), then they can be satisfied by *the documents received from Nantes Habitat Agency* ($q70$) with a support of 15.2% and a confidence of 82.9%.

$R_1$:  $q2=1$ $q3=1$ $q47=1$ ==> $q70=1$
Support = 15.2%   Confidence = 85.9%

### 4.2. Ontology and Ontology-Database Mapping

Ontology is defined basically by two main elements: a set of concepts ($C$) hierarchized by the subsumption relation and a set of relations ($R$) over concepts.

**4.2.1. Conceptual structure of the ontology.** We propose an ontology composed of two main parts, as shown in Figure 2.

The first one is a database items organisation with the root defined by the *Attributes* concept. The items are organized among the thematically structure of questions in the *Nantes Habitat* questionnaire. For instance, considering the *District* concept; it regroups fourteen questions (from $q1$ to $q14$) concerning the facilities and the quality of life in the district.

---

[2]  http://www.cs.waikato.ac.nz/ml/weka/

The second hierarchy, *Topic*, regroups all concepts created using necessary and sufficient conditions over other concepts.

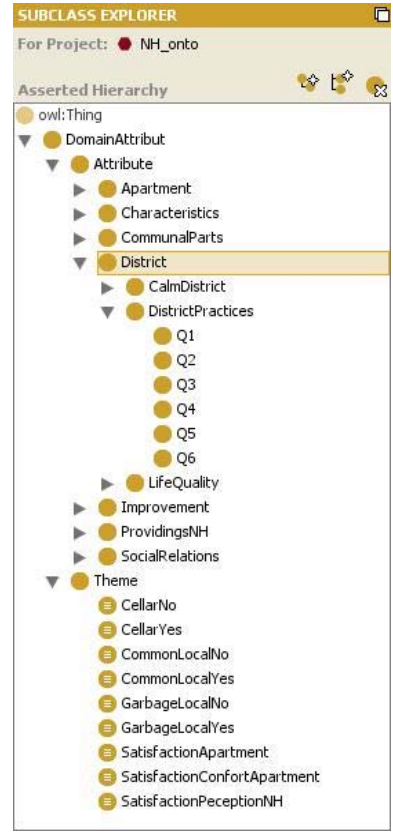

**Figure 2. Ontology structure in Protégé**[3]

For instance, considering the concept *SatisfactionDistrict*; in natural language, it describes the questions concerning the district with a satisfied answer. In other words, an item belongs to *SatisfactionDistrict* concept if it represents a question between $q1$ and $q14$, subsumed by the *District* concept, with a satisfied answer (*1* or *2*).

Moreover, the subsumption relation ($\leq$) is completed by the relation *hasAnswer* associating the *Attributes* concepts to an integer, simulating the relation attribute-value in the database.

To describe the ontology we use the Web Semantic representation language, OWL-DL[4]. Based on description logics, OWL-DL language permits, along with the ontological structure, to create concepts using necessary and sufficient conditions over other concepts. Also, we use the Protégé software to edit the ontology.

---

[3]  http://protege.stanford.edu/
[4]  http://www.w3.org/TR/owl-features

**4.2.2. Ontology-Database Mapping.** Part of rule schema definition, ontology concepts are mapped to a/several items in the database. Thus, several ontology-database connection types can be conceived.

Firstly, the simplest ontology-database mapping is the direct one. It connects one leaf-concept of the *Attribute* hierarchy to a set of items (semantically, the nearest one).

Considering the concept *Q1* of the ontology; it is associated to the attribute *q1="Are you satisfied with the transport in your district?"*. Furthermore, the concept *Q1* is instantiated in the ontology by 6 instances describing question *Q1* with 6 possible answers. Thus, the concept *Q1* is connected to 6 items as following:

$$f_0(Q1) = \{q1=1, q1=2, q1=3, q1=4, q1=95, q1=99\}$$

A second type of connection implies connecting concepts of *Topic* hierarchy to database. Considering the concept *SatisfComfortApartment* (Figure 2); in natural language, it is defined as all the concepts, subsumed by *ComfortApartment* (connected to questions *q44* to *q48*) and with a satisfied answer.

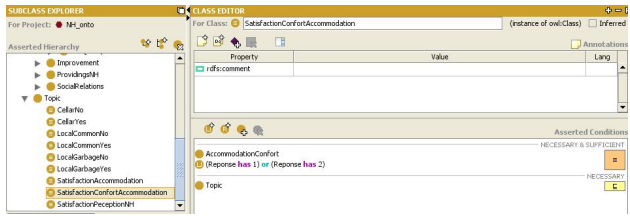

**Figure. 2. Concept construction using necessary and sufficient conditions in Protégé**

This definition over concepts can be translated in one definition over the set of items *I* by the following logical expression:

$C = SatisfComfortApartment$

$f(C) = \{ i_j \mid i_j \leq ConfortApartment \wedge$

$[ hasAnswer(i_j, 1) \text{ or } hasAnswer(i_j, 2) ] \} =$

$= \{q44=1, ..., q48=1, q44=2, ..., q48=2\}$

### 4.3. Rule Schemas

A rule schema allows user expectations representation and permits to the user to supervise association rule mining, meanwhile operators guide the post-processing task by pruning and filtering discovered rules. For example, let us consider the set of rule schemas with the operators presented in Table 2.

**Table 2. Operators et Rule Schemas**

| Rule Schema | Operator |
|---|---|
| $RS_1$:<SatFirstAppearance→BuildingsCondition> | $P(RS_1)$ |
| $RS_2$: <SatGarbagePlace→SatCommonPlace> | $P(RS_2)$ |
| $RS_3$: <UnsatPrice, UnsatCalmDistrict> | $C(RS_3)$ |
| $RS_4$: <SatComfortApartment→SatDelais> | $C(RS_4)$ |
| $RS_5$:<UnsatComfortApartment → UnsatHostListen> | $U(RS_5)$ |

The third rule, a non implicative rule schema, expresses possible relationship, without knowing the direction of the association, between the dissatisfaction concerning the price and the dissatisfaction concerning the quietness of the district. The first rule exemplifies an implicative rule schema representing the precise association between the satisfaction concerning the first appearance of the building and the one concerning the condition of the building.

Moreover, three of four operators proposed are developed: pruning operator is applied over the rule schemas $RS_1$ and $RS_2$ removing the two rule families, conforming operator is applied over the rules $RS_3$ and $RS_4$ and unexpectedness operator is exemplified using rule schema $RS_5$.

For instance, let us consider the fifth rule schema. To describe the concepts present in the definition we use the domain ontology as already exposed in the previous section. Thus, the concepts *UnsatComfortApartment* ($C_{d1}$) and *UnsatHostListen* ($C_{d2}$) are associated to a set of items as follows:

$f(C_{d1}) = \{q44=3, ..., q48=3, q44=4, ..., q48=4\}$

$f(C_{d2}) = \{q63=3, q73=3, q63=4, q73=4\}$

The developed framework permits an iterative process of pruning and filtering discovered rules. As a first action, the user uses the pruning operator applied over the first two rule schemas. The operator $P(RS_1)$ eliminates 1,974 association rules, meanwhile the operator $P(RS_2)$ removes 8,743 rules. The total of pruned rules is above 10% of the 82,159 discovered rules.

The conforming operator is applied over rules schemas $RS_3$ and $RS_4$. Thus the operator $C(RS_3)$ filters 7 association rules over the remaining of 70,000 rules. The operator $C(RS_4)$ filters 1,024 of the discovered rules.

The last operator, $U(RS_5)$, extracts 4 rules (Table 3) unexpected with respect to the condition. For example, let us verify if the following association rule extracted by *Apriori* is filter by this operator:

$RA_1$:  q58=4 q59=4 q62=4 => q63=4
    Support = 1.9%   Confidence = 81.5%

The rule $RA_1$ is unexpected with respect to the condition of the rule schema $RS_5$ if there is no item connected to the concept *UnsatComfortApartment* in the condition of the rule $RA_1$. Respectively, at least one item connected to the concept *UnsatHostListen* should be in the conclusion of the rule $RA_1$.

**Table 3. Association rules filtering**

| Antecedent | Consequent | Confidence | Support |
|---|---|---|---|
| q62=4,q64=4 | q63=4 | 0.852 | 0.019 |
| q64=4,q97=4 | q73=4 | 0.805 | 0.019 |
| q62=4,q72=4 | q63=4 | 0.815 | 0.020 |
| q58=4,q59=4,q62=4 | q63=4 | 0.815 | 0.019 |

Verifying the description of the two concepts in the ontology and comparing them with the items in the rule $RA_1$, we can observe that $i_1=$"q63=4" is an item connected to the concept *UnsatHostListen* because $i_1 \in f(C_{d2})$ and that, respectively, no item of $RA_1$ is connected to the concept *UnsatComfortApartment*. Thus, the rule $RA_1$ is unexpected with respect to the condition to rule schema $RS_5$.

## 5. Conclusion

This paper discusses the problem of helping the decision maker in the post-processing step of association rule mining. We propose to prune and filter discovered rule integrating user knowledge and beliefs.

User knowledge is modelled in an ontology connected to data. Rule schemas allow user belief representation, and, combined with ontologies, they improve the selection of interesting rules.

We intend to improve this approach in two directions:
- Developing the rule schema formalism;
- Integrating the approach in the discovery algorithm.